\documentclass[conference]{IEEEtran}
\usepackage{times}

\usepackage[numbers]{natbib}
\usepackage{multicol}

\usepackage[bookmarks=true,hidelinks]{hyperref}

\usepackage{graphicx,psfrag,color} 
  \graphicspath{ {./images/} }
\usepackage{amsmath}               
  \allowdisplaybreaks[1]           
\usepackage{amssymb}               
\usepackage{url}                   
\usepackage{rotating}              
\usepackage{multirow}              
\usepackage{lscape}                
\usepackage{tabularx}              
\usepackage{verbatim}              
\usepackage{footnote}              
\usepackage{float}                 
\usepackage{booktabs}              
\usepackage{lipsum}                
\usepackage{subcaption}            
\usepackage{indentfirst}
\usepackage[ruled]{algorithm2e} %
\usepackage{xcolor}
\usepackage{physics}

\usepackage{nicematrix} 
\usepackage{mathtools}

\usepackage{color}
\usepackage{pifont}
\usepackage{adjustbox}
\usepackage{romannum}
\usepackage{diagbox}
\usepackage{bm}

\usepackage{url}

\usepackage[normalem]{ulem}

\usepackage{color}
\definecolor{amethyst}{rgb}{0.6, 0.4, 0.8}

\newcommand{\rev}[2]{#2}

\begin{document}


\title{Function Based Sim-to-Real Learning for Shape Control of Deformable Free-form Surfaces}









%
\author{\authorblockN{Yingjun Tian\authorrefmark{2},
Guoxin Fang\authorrefmark{3},
Renbo Su\authorrefmark{2}, 
Weiming Wang\authorrefmark{2},
Simeon Gill\authorrefmark{2}, \\
Andrew Weightman\authorrefmark{2} and 
Charlie C.L. Wang\authorrefmark{2}\authorrefmark{1}} \\

\authorblockA{\authorrefmark{2}The University of Manchester}
\authorblockA{\authorrefmark{3}The Chinese University of Hong Kong}
\authorblockA{(\authorrefmark{1}Corresponding Author; 
Email:~changling.wang@manchester.ac.uk)}


}

\maketitle

\bstctlcite{IEEEexample:BSTcontrol}

\begin{abstract}
For the shape control of deformable free-form surfaces, simulation plays a crucial role in establishing the mapping between the actuation parameters and the deformed shapes. The differentiation of this forward kinematic mapping is usually employed to solve the inverse kinematic problem for determining the actuation parameters that can realize a target shape. However, the free-form surfaces obtained from simulators are always different from the physically deformed shapes due to the errors introduced by hardware and the simplification adopted in physical simulation. To fill the gap, we propose a novel deformation function based sim-to-real learning method that can map the geometric shape of a simulated model into its corresponding shape of the physical model. Unlike the existing sim-to-real learning methods that rely on completely acquired dense markers, our method accommodates sparsely distributed markers and can resiliently use all captured frames -- even for those in the presence of missing markers. To demonstrate its effectiveness, our sim-to-real method has been integrated into a neural network-based computational pipeline designed to tackle the inverse kinematic problem on a pneumatically actuated deformable mannequin.
\end{abstract}

\IEEEpeerreviewmaketitle

\thispagestyle{plain}
\pagestyle{plain}

\pagenumbering{arabic}

\section{Introduction}\label{secIntro}
Soft robotic systems have demonstrated the ability of dynamically adapting their surfaces in response to programmed actuation or environmental stimuli. Existing prototypes were mainly developed for haptic interfaces (ref.~\cite{inForm2013UIST,Stanley2015CtrlSurf,Bryan2019SoRoInterface,koehler2020model,Je2021ElevateAW}). 
\begin{figure}[!t] 
\centering
\includegraphics[width=\linewidth]{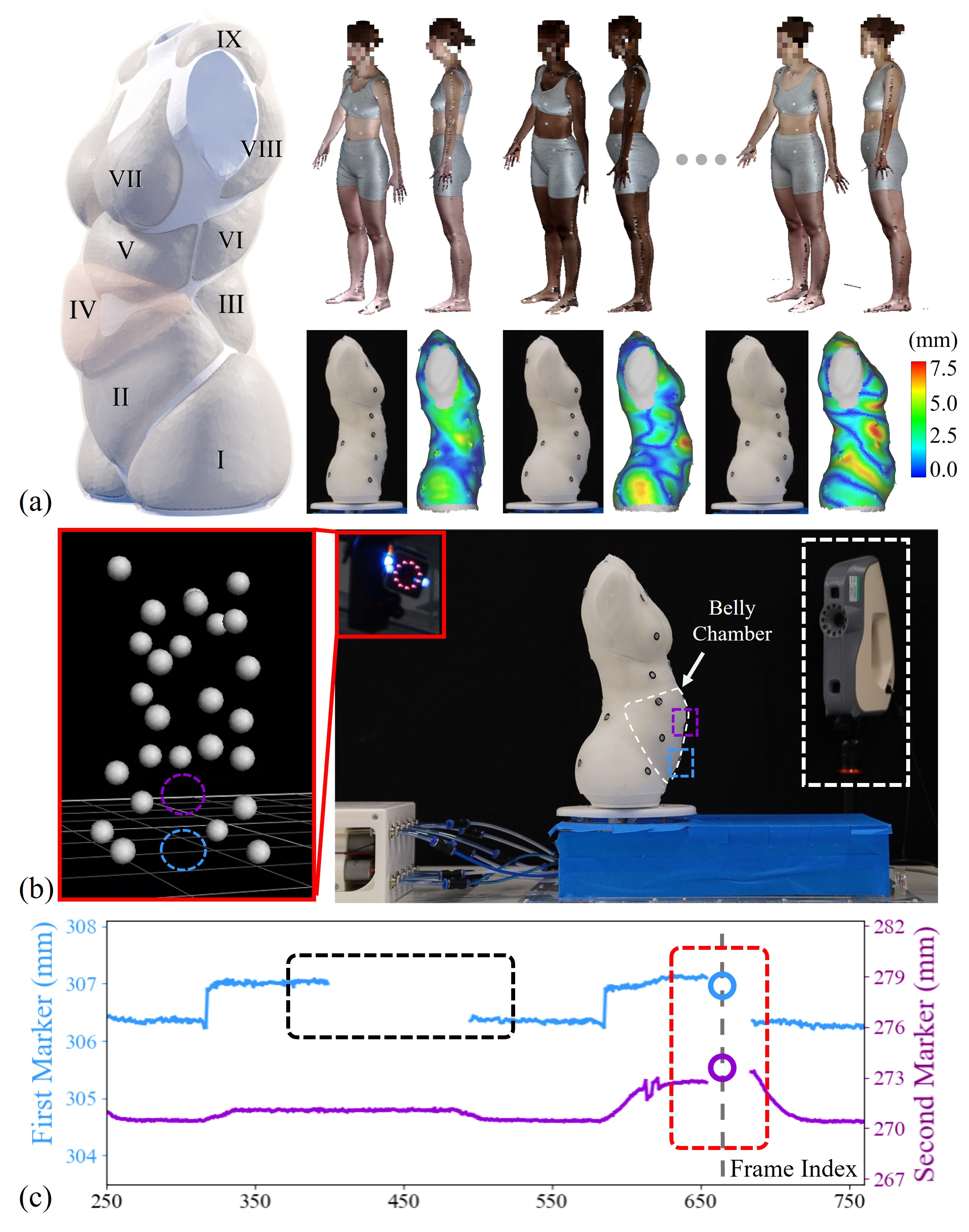}
\caption{Motivation and challenges. (a) A soft robotic mannequin with nine air chambers can be deformed by pneumatic actuation to realize the target body shape of individual customers. Shape approximation errors are visualized by colors. (b) The hardware setup includes a 3D scanner and a motion capture system. The zoom-in view shows the captured markers, where two markers on the belly were missed. (c) Positions of two markers captured while deforming the mannequin. 
}\label{fig:motivation}
\end{figure}
For the fabrication of customized garments, a soft robotic mannequin with deformable free-form surface has been developed in~\cite{Tian2022SoRoMannequin}, which is driven by pneumatic actuators to mimic the front shapes of different human bodies on the same mannequin. The research work presented in this paper is to realize a similar function of soft deformable robot\rev{}{, which is to serve as a reusable mould with programmable shape for sustainable garment fabrication (e.g.,~\cite{morby2016,parkes_2022})}.
Specifically, by giving a target shape of an individual as $\mathcal{S}^t$, the deformable mannequin is expected to deform into a shape $\mathcal{S}$ that minimizes the shape difference (see Fig.\ref{fig:motivation}(a) for examples). The geometric errors are evaluated by capturing the deformed shapes of the robotic mannequin with the help of a structure-light based 3D scanner (see Fig.\ref{fig:motivation}(b)). Shape control of a deformable free-form surface is challenging, and the reasons for this are analyzed below with a review of related work in the literature.


\subsection{Kinematics for Shape Control}\label{subsecIntroShapeCtrl}
An essential shape control problem is to compute the actuation (denoted by a vector $\mathbf{a}$) that can drive the robotic free-form surface $\mathcal{S}$ to approach a target shape $\mathcal{S}^t$. This is an inverse computing problem, and the shape control of free-form surfaces mainly focused on the configuration space (e.g.,~\cite{george2018control}). It presents much higher \textit{degrees-of-freedom} (DoFs) compared to the \textit{inverse kinematics} (IK) problems for articulated robots that generally focus on the task space. This leads to high complexity in both the forward shape estimation (i.e., predict the shape $\mathcal{S}(\mathbf{a})$) and the IK computation to determine the value of $\mathbf{a}$. An intuitive solution is to \rev{directly capture shape data from the physical setup and apply a Jacobian-based iterative solver for IK}{apply an iteration-based IK solver with Jacobian physically evaluated on hardware setup}~\cite{Tian2022SoRoMannequin, Yip2014TROHardwardGradient}. However, this approach suffers from high costs in both computation and data acquisition, and its performance heavily relies on the robustness of data acquisition and the accuracy of shapes captured on the physical setup. 


In the existing research works of kinematics computing for deformable objects, the high DoFs in the configuration space are typically modeled either by reduced analytical models or by discrete numerical models. \rev{}{Comprehensive reviews can be found in the papers of Armanini \textit{et al.}~\cite{Renda2023TROModeling} and Arriola-Rios \textit{et al.}~\cite{arriola2020modeling}.} Numerical simulation is commonly employed as a general solution for modeling deformable objects with complex shapes, and it can be adapted in both the cases with internal actuation~\cite{Guoxin2020TRO} and those with external interactions~\cite{ficuciello2018fem}. However, high-precision numerical simulation in general cannot be computed efficiently enough to realize a fast IK solver. Moreover, numerical simulation invariably exhibits a prediction error compared to physical performance, a discrepancy often referred to as the `reality gap'~\cite{rss2020Sim2Real}.

A widely adopted strategy to mitigate the issue of efficiency is to first collect a dataset using the numerical solution and train a \textit{neural network} (NN) for the forward shape estimation of the deformable object~\cite{Thomas2018SoRoCtrlReview, bern2020soft}. Subsequently, NN-based sim-to-real transfer~\cite{Dubied2022Sim2RealFEM, Zhang2022Sim2RealSoRoFish, Iyengar2023Sim2RealCTR, Fang2022Sim2Real} is applied to achieve high precision while maintaining reasonable efficiency in data acquisition -- i.e., only a small set of physical data is needed. When both the forward kinematic computing and the sim-to-real transfer are realized by neural networks, the gradients of IK objectives can be efficiently and effectively computed by differentiating the networks~\cite{du2021_diffpd, hu2019chainqueen}. In short, fast IK can be achieved. 

\subsection{Challenges of Sim-to-Real Transfer}\label{subsecIntroSim2Real}
%

%
Sim-to-real approaches have been developed for controlling robots with compliance~\cite{Iyengar2023Sim2RealCTR,shah2021soft} and for deformable object manipulation~\cite{Fei2021ICRASim2Real,liang2023real}. These works generally adopt rich physical information obtained from well-captured frames with constant DOFs as training data. Less effort has been focused on the sim-to-real problem on free-form surfaces using sparse and incomplete sets of markers. 
There is an effort to use a 3D scanner to directly obtain the surface shapes as point clouds~\cite{Forte2022InverseDesignMembrane}, which however suffers from the problems of inefficiency and unstable boundary regions. Moreover, the scanned shapes at different time currents have no correspondence with each other, which needs a lot of extra effort for non-rigid registration~\cite{deng2022survey}.

Using a motion capture system can precisely capture the position of markers attached to a free-form surface~\cite{Mehrnoosh2022MarkerSim2Real, scharff2021sensing}. 
However, only sparse data can be obtained, representing a reduced order of information on the shape of the free-form surface. The motion capture system can also suffer from the problems of marker missing due to camera view occlusion -- especially when large deformation is presented (see Fig.\ref{fig:motivation}(b, c) for an example). In this paper, we propose a function-based learning method to address the sim-to-real challenge, and thereby also solve the fast IK problem for deformable free-form surfaces.

\subsection{Our Method}\label{subsecIntroShapeCtrl}


\begin{table*}[t]
\footnotesize
\centering
\caption{\rev{}{List of Notations}}\label{tab:Symbols}
\begin{tabular}{cl|cl}
\hline \hline
\specialrule{0em}{2pt}{1pt}
Symbol & Description  & Symbol & Description  \\ 
\specialrule{0em}{1pt}{1pt} \hline \specialrule{0em}{1pt}{2pt}

$ \mathbf{a} $  & Actuation parameters (i.e., chamber pressures)
&  $\mathcal{S}^t$  & The target body shape to be achieved by IK \\

$\mathcal{S}$ & The simulated shape of the deformable surface & 
 $\mathcal{S}^*$  & The fixed shape after applying our \textit{sim-to-real} method \\

$\mathcal{S}^c$ & The compact shape descriptor (i.e., control points) of $\mathcal{S}$ &
 $\mathcal{S}^p$    & The scanned 3D shape on the physical setup \\

 $\mathcal{N}_{fk}$ & Forward kinematics network that predicts $\mathcal{S}^c$ from $ \mathbf{a} $ & $\mathcal{N}_{mk}$& The baseline network predicting positions of markers \\

 $\mathcal{N}_{rbf}$ & Function prediction network that predicts $\bm{\gamma}$ by $\mathcal{S}^c$ &  $\mathcal{N}_{s2r}$& Combination of $\mathcal{N}_{rbf}$ and $\bm{\Phi}(\cdot)$  \\

$\bm{\Phi}(\cdot)$ & The space warping function used to deform $\mathcal{S}$ to $\mathcal{S}^*$ & 
$\bm{\gamma} = [\bm{\alpha}_0, \bm{\alpha}_1, ...  \bm{\beta}_i]$ & Variables to determine the function space
\\ 

$\{ \mathbf{q}_i \}$& Centers of Gaussian kernels (also markers on $\mathcal{S}$) in $\bm{\Phi}(\mathbf{p})$ & $\mathbf{p}$, $\mathbf{p}^* \in \mathbb{R}^{3}$ & The surface point on $\mathcal{S}$ and $\mathcal{S}^*$ respectively \\

$(\mathbf{R},\mathbf{t})$ & Rotation matrix and translation vector to re-pose $\mathcal{S}^t$&
$\mathbf{c}^*$  & Closet point of $\mathbf{p}^*$ on $\mathcal{S}^t$ \\


$(u_p,v_p)$ & A point in the parametric domain of B-spline surface &
$\mathbf{B}(\cdot)$ & B-Spline function as decoder mapping $(u_p,v_p)$ to $\mathbf{p}$\\

\specialrule{0em}{1pt}{1pt} \hline\hline
\end{tabular}\label{table_variable}
\end{table*}

To facilitate the neural network-based fast IK computing, we propose a resilient sim-to-real learning method, which can fully leverage the entire dataset of sparse marker positions obtained from the physical setup. It is a new learning architecture that learns the function space of shape deformation represented by \textit{radial basis functions} (RBF). Specifically, each simulated free-form surface is approximated by a B-spline surface, the control points of which are employed as the shape descriptor in a lower dimension. The input of our sim-to-real network is the control points of a simulated body shape, and the output is the coefficients of RBFs that define a spatial warping function $\bm{\Phi}(\mathbf{p})$ by using the positions of `virtual' markers as kernel centers. The sim-to-real network can be trained from sparse and even incompletely captured sets of markers.

The space warping function $\bm{\Phi}(\mathbf{p})$ is in a closed-form and gives a continuous mapping from the shape of a simulated model to the shape of a fixed model that matches the reality. Working together with the forward kinematics prediction that is trained by simulation, this forms an end-to-end NN-based pipeline that can precisely predict the physical shapes of robotic deformable surfaces. This computation pipeline is analytically differentiable so that the gradient-based iteration for IK can be computed efficiently. The effectiveness of our fast IK solver has been verified on a physical setup as shown in Fig.\ref{fig:motivation}(b). When testing on a dataset with $100$ individual shapes, our IK solver takes about $7.98$ seconds in average to generate the actuation parameters. The quality of IK solutions has been verified by scanning the physical shape $\mathcal{S}^p$ to compare with the input target shape $\mathcal{S}^t$. The resultant error color maps for different individuals are shown in Fig.\ref{fig:motivation}(a). 


The technique contributions of our work are summarized as follows:
\begin{itemize}
\item A resilient sim-to-real learning method for deformable free-form surfaces that can employ sparse and incomplete sets of markers to bridge the reality gap by RBF-based spatial warping functions;

\item A neural network-based method that can efficiently solve the inverse kinematics problem with its performance physically verified on a pneumatically actuated deformable mannequin. 
\end{itemize}
Experimental tests were conducted on a variety of individual shapes presented in the CAESAR dataset~\cite{CAESAR2002}. The shape approximation errors were physically evaluated on a deformable mannequin with 1:2 ratio to the full size. 
Detail statistics are given in the supplementary document.
\section{Learning for Sim-to-Real Transfer}\label{secSim2Real}
This section introduces our function based sim-to-real learning for shape prediction. Given a set of points $\{ \mathbf{p} \in \mathbb{R}^3 \}$ on the free-form surface of a simulated model $\mathcal{S}$, we propose a method to learn a network $\mathcal{N}_{s2r}$ that can predict a deformation function mapping these points onto a fixed model $\mathcal{S}^*$ (\rev{}{i.e., the  fixed shape that has eliminated the sim-to-real gap). All notations used in this paper are summarized in Table \ref{tab:Symbols}.} The shape difference between $\mathcal{S}^*$ and the physical shape $\mathcal{S}^p$ should be minimized by the sim-to-real transfer. $\mathcal{N}_{s2r}$ is expected to be a general network that can adaptively handle different simulated shapes as input.

\subsection{RBF-based Spatial Warping}\label{subsecRBFSpaceWarping}
Considering the gap between the simulated shape $\mathcal{S}$ and the physical shape $\mathcal{S}^p$, we propose to model it as a continuous spatial warping by the \textit{radial basis functions} (RBF) as follows. 
\begin{equation}\label{eqRBFSpaceWarping}
\mathbf{p}^* = \bm{\Phi}(\mathbf{p}) = \bm{\alpha}_0 + A \mathbf{p} + \sum_{i=1}^N \bm{\beta}_i e^{-c\| \mathbf{p} - \mathbf{q}_i \|^2} 
\end{equation}
with $N$ being the number of kernels employed for the space warping and $A = [\bm{\alpha}_1, \bm{\alpha}_2, \bm{\alpha}_3]$. The centers of Gaussian kernels, $\{ \mathbf{q}_i \}$, are sparsely selected to cover the entire surface of a simulated model. The value of coefficient $c$ controlling the width of the Gaussian kernel is chosen as $(3.0e-5)$ by experiments. In our practice, we use the locations of markers for motion capture as the kernel centers. We remark that $\bm{\gamma} = [\bm{\alpha}_0, \bm{\alpha}_1, ...  \bm{\beta}_i]$ are variables of function space, which can be determined by solving a linear system \cite{Charlie2007RBFVolPara}. Note that $\{ \mathbf{q}_i\}$ are virtual markers defined on the simulated models, where the free-form surface of a simulated model has been parameterized into $u,v$-domain and represented by a B-spline surface. The same marker on different simulated models will have the same $(u, v)$ parameters.

\begin{figure}[t] 
\centering
\includegraphics[width=1.0\linewidth]{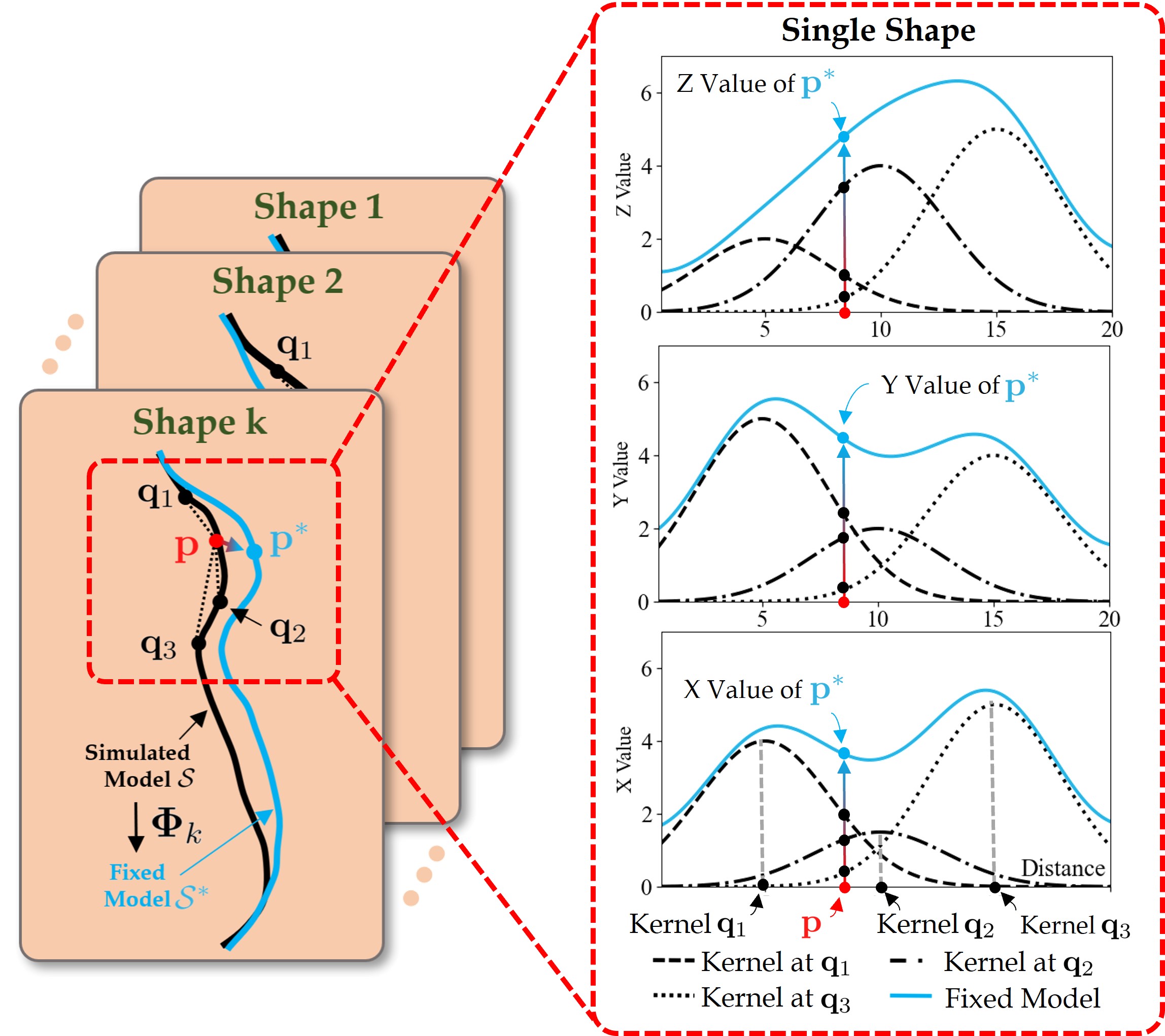}\\
\vspace{-2pt}
\caption{ 
The space warping $\mathbf{\Phi}(\cdot)$ is built on the \textit{radial basis functions} (RBF) with kernel centers located on the free-form surface of a simulated model. Different warping functions need to be determined for different simulated shapes. 
}\label{fig:spaceWarping}
\end{figure}

\begin{figure}
\centering
\includegraphics[width=1.0\linewidth]{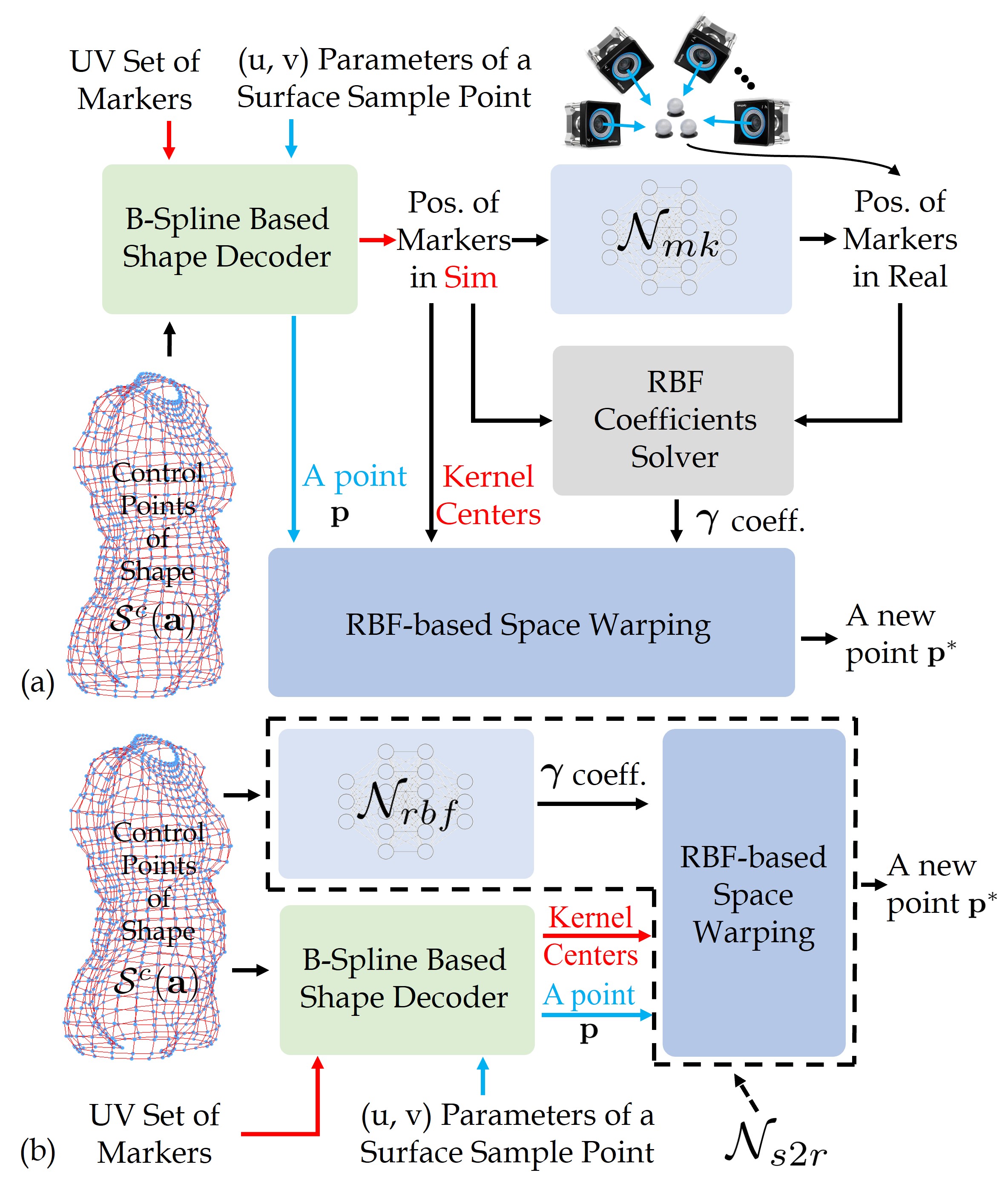}
\caption{Different sim-to-real strategies by using RBF-based spatial warping -- (a) a marker-prediction based pipeline (\rev{}{the baseline method that can only use frames with complete set of markers}) that needs to solve different linear systems for different simulated shapes and (b) our function-prediction based end-to-end pipeline (\rev{}{that can resiliently use all frames -- even for those in the presence of missing markers}) where the components circled by dash lines form a sim-to-real network $\mathcal{N}_{s2r}$.}
\label{fig:comparisonSim2RealPipelines}
\end{figure}

With a determined warping function $\bm{\Phi}(\mathbf{p})$, we are able to map any point $\mathbf{p} \in \mathcal{S}$ to its corresponding point $\mathbf{p}^* \in \mathcal{S}^*$ as demonstrated in Fig.\ref{fig:spaceWarping}. However, different warping functions need to be employed for different simulated shapes. A straightforward solution is to train a network $\mathcal{N}_{mk}$ for predicting the positions of physical markers by using the set of virtual marker positions as $\{ \mathbf{q}^* \}=\mathcal{N}_{mk}(\{ \mathbf{q} \})$. As a result, the predicted positions of physical markers and the positions of virtual markers can be used to determine the value of $\bm{\gamma}$ therefore also the warping function -- see Fig.\ref{fig:comparisonSim2RealPipelines}(a) for an illustration of this marker-prediction based pipeline. There are two major problems with this sim-to-real strategy:
\begin{itemize}
\item This is not an end-to-end approach as it needs to solve a linear equation system when establishing a warping function for each simulated shape;

\item Training the marker prediction network $\mathcal{N}_{mk}$ needs sets of fully captured markers, which can be challenging on largely deformed models as discussed above.
\end{itemize}
A function space learning-based method is introduced below to solve these issues.

\subsection{NN-based Deformation Function Learning}\label{subsecNNBasedDefField}
Only learning to predicting the positions of markers does not provide good shape-control in the regions between markers, which relies on dense and complete sets of captured markers as input to train $\mathcal{N}_{mk}$. Differently, we propose a function-prediction based pipeline, which is end-to-end after training. Moreover, all frames of motion capture data -- even those in the presence of missing markers can be employed for training. 

A network $\mathcal{N}_{rbf}$ is introduced to predict the value of the function variable $\bm{\gamma}$ according to the simulated shape $\mathcal{S}$. However, using all sample points on the surface of $\mathcal{S}$ may bring in redundant information, and therefore require a complex network for the prediction. We employ the control points of B-spline surface (denoted by $\mathcal{S}^c$) as the shape descriptor of $\mathcal{S}$, with which any point in the parametric domain with the parameter $(u_p,v_p)$ can be mapped to a point $\mathbf{p} \in \mathbb{R}^3$ as $\mathbf{p}=\mathbf{B}(u_p,v_p,\mathcal{S}^c)$. This function $\mathbf{B}(\cdot)$ is named as the B-spline based shape decoder in our computational pipeline (see the green block in Fig.\ref{fig:comparisonSim2RealPipelines} for an example). 

Our function-prediction based sim-to-real pipeline is formed by three building blocks: 1) the network $\mathcal{N}_{rbf}$, 2) the B-spline based shape decoder and 3) the RBF-based space warping (i.e., Eq.(\ref{eqRBFSpaceWarping})), where the diagram has been shown in Fig.\ref{fig:comparisonSim2RealPipelines}(b). When actuating the deformable mannequin into different shapes $\{ \mathcal{S}_j \}$ by $\{ \mathbf{a}_j \}$, the function-prediction network $\mathcal{N}_{rbf}$ can be trained by the position of all captured markers $\mathbf{q}_{i,j}$ together with their corresponding parameters $(u_i,v_i)$ and the control points $\mathcal{S}^c_j$. Here $\mathbf{q}_{i,j}$ denotes the position of the $i$-th marker captured on the $j$-th frame. The loss function for learning $\mathcal{N}_{rbf}(\cdot)$ is defined as 
\begin{equation}\label{eqS2RLoss}
    \mathcal{L}_{s2r} = \sum_j \sum_i \| \bm{\Phi}_j(\mathbf{B}(u_i,v_i,\mathcal{S}^c_j)) - \mathbf{q}_{i,j} \|^2
\end{equation}
with the function variable $\bm{\gamma}$ of $\bm{\Phi}_j$ predicted by the network as $\bm{\gamma}=\mathcal{N}_{rbf}(\mathcal{S}^c_j)$ and the kernel centers of $\bm{\Phi}_j$ given as $\{ \mathbf{B}(u_i,v_i,\mathcal{S}^c_j) \}$. Only captured markers are included to evaluate the loss $\mathcal{L}_{s2r}$ defined in Eq.(\ref{eqS2RLoss}). This pipeline supports resilient sim-to-real learning that can use all frames of captured data by motion camera. 

Given a well trained network $\mathcal{N}_{rbf}(\cdot)$, the sim-to-real transfer can be evaluated by the pipeline of Fig.\ref{fig:comparisonSim2RealPipelines}(b) in the inference phase as a network $\mathcal{N}_{s2r}(\cdot)$ that gives
\begin{equation} \label{eqSim2RealPrediction}
\mathbf{p}^*=\mathcal{N}_{s2r}(\mathbf{B}(u_p,v_p,\mathcal{S}^c(\mathbf{a})),\mathcal{S}^c(\mathbf{a}),\{ \mathbf{B}(u_q,v_q,\mathcal{S}^c(\mathbf{a})\})
\end{equation}
$(u_p,v_p)$ denotes a point sampled in the parametric domain of the free-form surface.  $\{\mathbf{B}(u_q,v_q,\mathcal{S}^c(\mathbf{a})\}$ denotes the set of marker points, which are defined by their parameters $(u_q,v_q)$ and are chosen as kernel centers for the RBF-based warping function. 
All operators in this sim-to-real pipeline is differentiable w.r.t. the actuation parameter $\mathbf{a}$. In the following section, we present the detail of NN-based forward kinematics and the IK computing for shape control, which is based on minimizing the distance between $\mathbf{p}^*$ and its closest point on the target shape $\mathcal{S}^t$.

\section{Neural Networks Based Kinematic Computing}\label{secNNShapeCtrl}
This section presents the neural network-based computational pipeline for kinematics. Shape control of deformable free-form surfaces can be realized by our fast IK solution.

\subsection{NN-based Forward Kinematics}\label{subsecNNBasedFK}
While the recently developed fast simulator provides an efficient solution to predict the deformed shape of soft robots~\cite{Guoxin2022IROS}, applying it to estimate gradients of the IK objective by numerical differentiation remains time-consuming~\cite{Guoxin2020TRO}. Here we propose to employ an NN-based computational pipeline for forward kinematics. Unlike those work interested in the status of end-effectors in the task space (often in a very low dimension), the output of our network as free-form surfaces can have a very high dimension if the shapes are simply represented as sample points. It's very challenging to train such a network, as a massive number of simulation results are needed.

\begin{figure}
\centering
\captionsetup{skip=5pt}  
\includegraphics[width=1.0\linewidth]{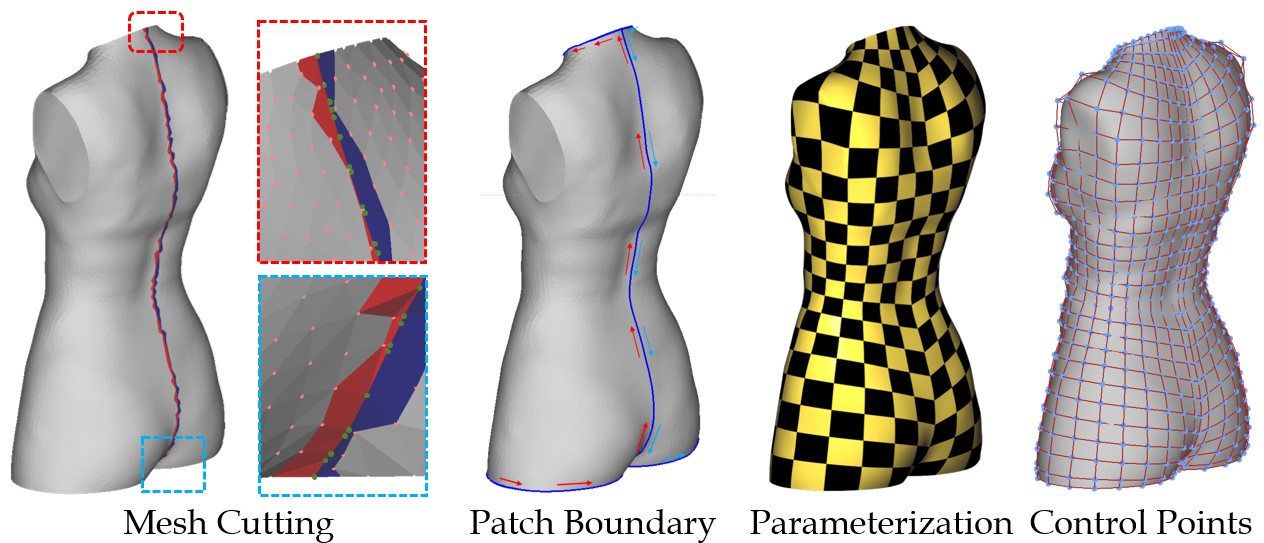}

\caption{Steps of generating the compact B-spline representation for the free-form surface of a deformable mannequin.}\label{fig:BSplineFitting}
\vspace{-10pt}
\end{figure}

\begin{figure}
\centering
\includegraphics[width=1.0\linewidth]{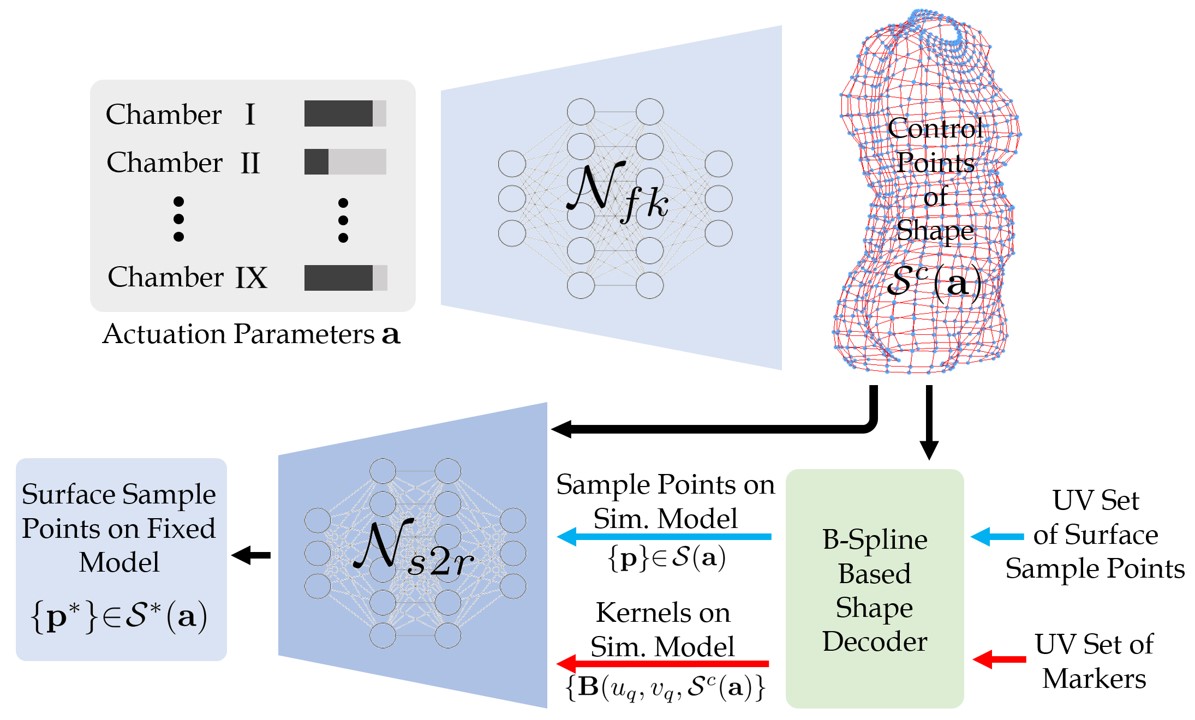}
\caption{NN-based pipeline for forward kinematic computing.}\label{fig:NNBasedFK}
\vspace{-15pt}
\end{figure}

In this work, we apply B-spline surface representation as a compact shape descriptor for each free-form surface $\mathcal{S}$ generated by simulation. Specifically, the triangular mesh of $\mathcal{S}$ can be approximated by a B-spline surface, with its control points $\mathcal{S}^c$ determined through parameterization and fitting steps (see Fig. \ref{fig:BSplineFitting} for an illustration). The dimension of $\mathcal{S}^c$ using B-spline representation is lower than mesh representation. In our implementation, we use $30 \times 30$ control points for a mannequin model as shown in Fig.\ref{fig:NNBasedFK}. Note that the parameterization is consistent among all models as they are generated by numerical simulation using the mesh representation with the same set of vertices and elements -- only the positions of vertices are changed.

Given a set of randomly generated actuation parameters as $\{ \mathbf{a}_k \}$, we can run the numerical simulators to obtain the deformed shapes $\{ \mathcal{S}^c \}$ and generate their corresponding control points as $\{ \mathcal{S}^c_k \}$. A training dataset with $M$ such pairs of results $\{ \mathbf{a}_k : \mathcal{S}^c_k \}_{k=1,\ldots,M}$ can be obtained by the simulator using a reduced physical model (i.e.,~\cite{Guoxin2022IROS}). It is important to generate a dataset that spans the entire space of shape variation. The dataset is then employed to train a network $\mathcal{N}_{fk}$ that can give the fast simulation result with an input actuation as $\mathcal{S}^c=\mathcal{N}_{fk}(\mathbf{a})$.

The gap between the simplified simulation and the physical reality can be well captured and fixed by our sim-to-real network. As a result, for any point $(u_p,u_p)$ sampled in the $u,v$-parametric domain, its position on the physical model can be predicted by 
\begin{equation}\label{eqSim2RealPrediction}
\mathbf{p}^*=\mathcal{N}_{s2r}(\mathbf{B}(u_p,v_p,\mathcal{N}_{fk}(\mathbf{a})),\mathcal{N}_{fk}(\mathbf{a}),\{ \mathbf{B}(u_q,v_q,\mathcal{N}_{fk}(\mathbf{a})\}).
\end{equation}
This NN-based computational pipeline for forward kinematics was illustrated in Fig.\ref{fig:NNBasedFK}. The predicted shape $\mathcal{S}^*(\mathbf{a})$ can be obtained as a set of points $\{ \mathbf{p}^* \}$, which is a differentiable function in terms of the actuation.

\subsection{Gradient Iteration Based Inverse Kinematics}\label{subsecGradientBasedIK}
The IK computation is formulated as an optimization problem that minimize the shape difference between the predicted shape $\mathcal{S}^*$ and the target shape $\mathcal{S}^t$. That is 

\begin{equation}
    \arg \min_{\mathbf{a},\mathbf{R},\mathbf{t}} D(\mathcal{S}^*(\mathbf{a}), \mathcal{S}^t) = 
        \sum_{\mathbf{p}^* \in \mathcal{S}^*(\mathbf{a})} \|\mathbf{p}^* -  \left(\mathbf{R} \mathbf{c}^* + \mathbf{t} \right) \|^2
\end{equation}
where the rotation matrix $\mathbf{R}$ and the translation vector $\mathbf{t}$ are applied to the target model to eliminate the influence of orientation variation. $\mathbf{c}^*$ is the closet point of $\mathbf{p}^*$ on the target shape $\mathcal{S}^t$. 

The objective function can be minimized using the gradient descent method with linear search. After each iteration of $\mathbf{a}$ update, we apply an ICP-based rigid registration to determine a new pair of $(\mathbf{R},\mathbf{t})$. In other words, the optimization process alternates between updating the actuation parameter $\mathbf{a}$ and adjusting the pose of target model $(\mathbf{R},\mathbf{t})$. Thanks to the formulation of NN-based forward kinematics and our function based sim-to-real learning, the gradients of $D(\cdot)$ can be computed analytically. That is
\begin{flalign}
    \frac{\partial D}{\partial \mathbf{a}} &=
    \sum_{\mathbf{p}^* \in \mathcal{S}^*(\mathbf{a})} 2 (\frac{\partial \mathbf{p}^*}{\partial \mathbf{a}})^{T}
   (\mathbf{p}^* -  \left(\mathbf{R} \mathbf{c}^* + \mathbf{t} \right)),
\end{flalign}
where 
\begin{flalign}
\frac{\partial \mathbf{p}^*}{\partial \mathbf{a}} &= \frac{\partial \mathcal{N}_{s2r}}{\partial \mathbf{B}} 
\frac{\partial \mathbf{B}}{\partial \mathbf{a}} + \frac{\partial \mathcal{N}_{s2r}}{\partial \mathcal{N}_{fk}}
\frac{\partial \mathcal{N}_{fk}}{\partial \mathbf{a}} +  \frac{\partial \mathcal{N}_{s2r}}{\partial \{\mathbf{B}\}} \frac{\partial \{\mathbf{B}\}}{\partial \mathbf{a}}
\nonumber
\\&= \left(\frac{\partial \mathcal{N}_{s2r}}{\partial \mathbf{B}} 
\frac{\partial \mathbf{B}} {\partial \mathcal{N}_{fk}}
+ \frac{\partial \mathcal{N}_{s2r}}{\partial \mathcal{N}_{fk}} + \frac{\partial \mathcal{N}_{s2r}}{\partial \{\mathbf{B}\}}  \frac{\partial \{\mathbf{B}\}} {\partial \mathcal{N}_{fk}}\right)\frac{\partial \mathcal{N}_{fk}}{\partial \mathbf{a}}.
\end{flalign}
Note that $\{\mathbf{B}\}$ denotes those terms contributed by the set of virtual markers (as kernels of RBF-base warping) while $ \mathbf{B}$ is for the surface point $\mathbf{p}$.

For realizing a fast iterative-based IK solver, the analytical evaluation of the gradient using network-based forward kinematics is crucial. It avoids the time-consuming and unstable steps of numerical difference taken in the simulation~\cite{Guoxin2020TRO} or on the physical setup~\cite{Tian2022SoRoMannequin, Yip2014TROHardwardGradient}.

\section{Implementation Details and Results}\label{secResult}
We have implemented the proposed approach in C++ and Python. The networks training part are implemented in the PyTouch platform with the batch size as $32$, the learning rate at $0.001$ and the maximum number of epochs as $150$. 
For the interference phase, we transferred the trained networks to be integrated with our IK solver in C++ running on CPU. The analytical gradients of the networks are employed in our IK solver. All the training and computational tests are conducted on a PC with Intel i7-12700H CPU, RTX $3060$ GPU and $32$ GB RAM. The results of sim-to-real and fast IK have been verified on a physical setup of soft deformable mannequin. 

\subsection{Details of Networks and Training}\label{subsecImplementationDetails}



For the training of forward kinematics network $\mathcal{N}_{fk}$, a dataset with 1,000 different pairs of actuation and shapes is collected from the simulation. 
\rev{}{This dataset contains shapes with the minimal and the maximal pressures applied to every chamber - in total 512 shapes (i.e., $2^9$ with 9 chambers) were generated. We also randomly sampled other 488 actuation parameters using the Halton sequence~\cite{halton1960efficiency}. All these 1,000 samples of actuation are applied in the simulation system to obtain their simulated shapes to explore the whole deformation space.} Considering the large non-linearity between actuation and shape, we non-uniformly resample the actuation space by the \rev{actual expansion ratio}{amount of chamber inflation} to enhance the training accuracy. The average time used for completing each simulation is around $150$ seconds. This dataset is separated in the ratio of $7:3$ for training and testing.

\begin{figure}[t] 
\centering
\includegraphics[width=1.0\linewidth]{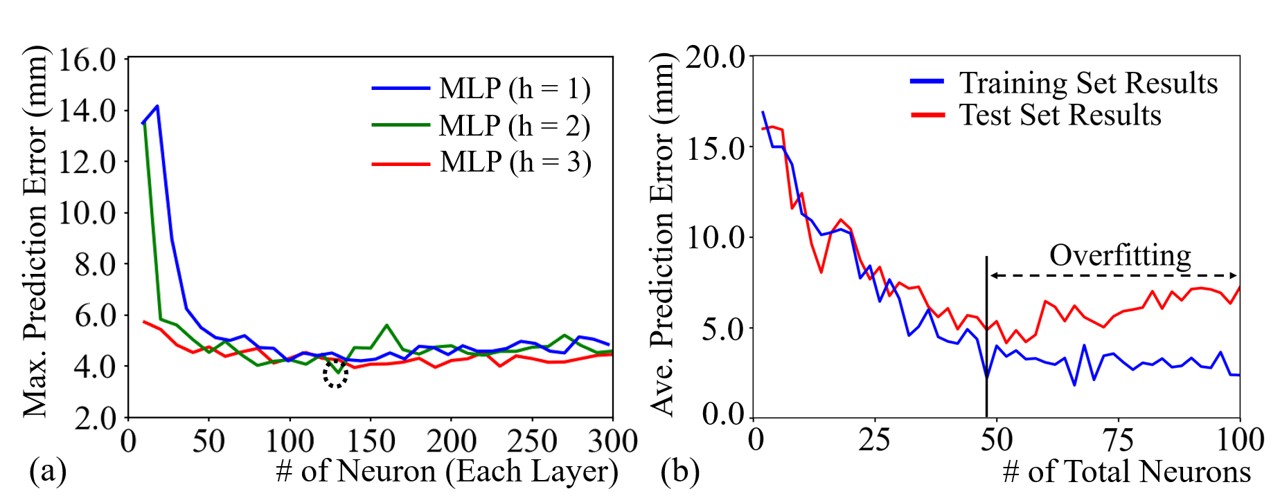}\\
\caption{\rev{}{Study of network parameters -- (a) the maximal shape prediction error w.r.t. different numbers of layers ($h$) and numbers of neurons per layer ($l$) are evaluated to choose the `best' values of $h$ and $l$ for $\mathcal{N}_{fk}$ (circled by black dash lines), and (b) the average shape prediction error w.r.t. the number of neurons in $\mathcal{N}_{rbf}$ is studied to avoid overfitting while training $\mathcal{N}_{s2r}$ using a limited number of samples. Note that $\mathcal{N}_{s2r}$ contains $\mathcal{N}_{rbf}$ plus an additional space warping module as illustrated in Fig.\ref{fig:comparisonSim2RealPipelines}.}}\label{fig:NfkTraining}
\end{figure}

We carefully select the network structure, which includes the number of hidden layers (\textit{h}), and the number of neurons in each hidden layer (\textit{l}). \rev{}{The experimental results of the maximal shape prediction error according to different network structures are studied and illustrated in Fig.~\ref{fig:NfkTraining}(a)}, and we choose two hidden layers with each layer contains 128 neurons as the final network structure to balance the \rev{}{quality of} training result and \rev{}{the} computational cost. \rev{}{In our implementation, ReLU is selected as the activation function and batch normalization is applied to improve stability in training.} Note that our experiment finds that prediction with better quality can be achieved when learning the translation vectors applied to the positions of control points on an average model of all training shapes.


For the function prediction network $\mathcal{N}_{rbf}$ of sim-to-real transfer, we employ a network architecture with 2 hidden layers where each layer \rev{has 24}{has a certain number of} neurons using the ReLU activation function. 
The input layer contains the positions of $30 \times 30$ control points, and the output layer is the coefficients for the RBF-based warping function as $\bm{\gamma} \in \mathbb{R}^{3(N+4)}$ with $N=34$ being the number of markers used in motion capture. The training of $\mathcal{N}_{rbf}$ is taken on 40 frames of markers obtained from the motion capture system while randomly varying the pneumatic actuation within the working range of the \rev{actuators}{chambers}. Among these 40 frames, around $18$ frames have missing markers mainly due to camera occlusion caused by large deformation on the physical mannequin. However, all captured markers (even for those from frames with missing markers) are employed as samples to train $\mathcal{N}_{rbf}$. In total, $1330$ points are involved in the training. 
\rev{}{By using this dataset, it can be observed from the study as shown in Fig.~\ref{fig:NfkTraining}(b) that the prediction error starts to increase when the total number of neurons exceeds 50 -- i.e., overfitting happens. Based on this analysis, we select 24 neurons for each layer in $\mathcal{N}_{rbf}$.
}

\begin{figure}[t] 
\centering
\includegraphics[width=1.0\linewidth]{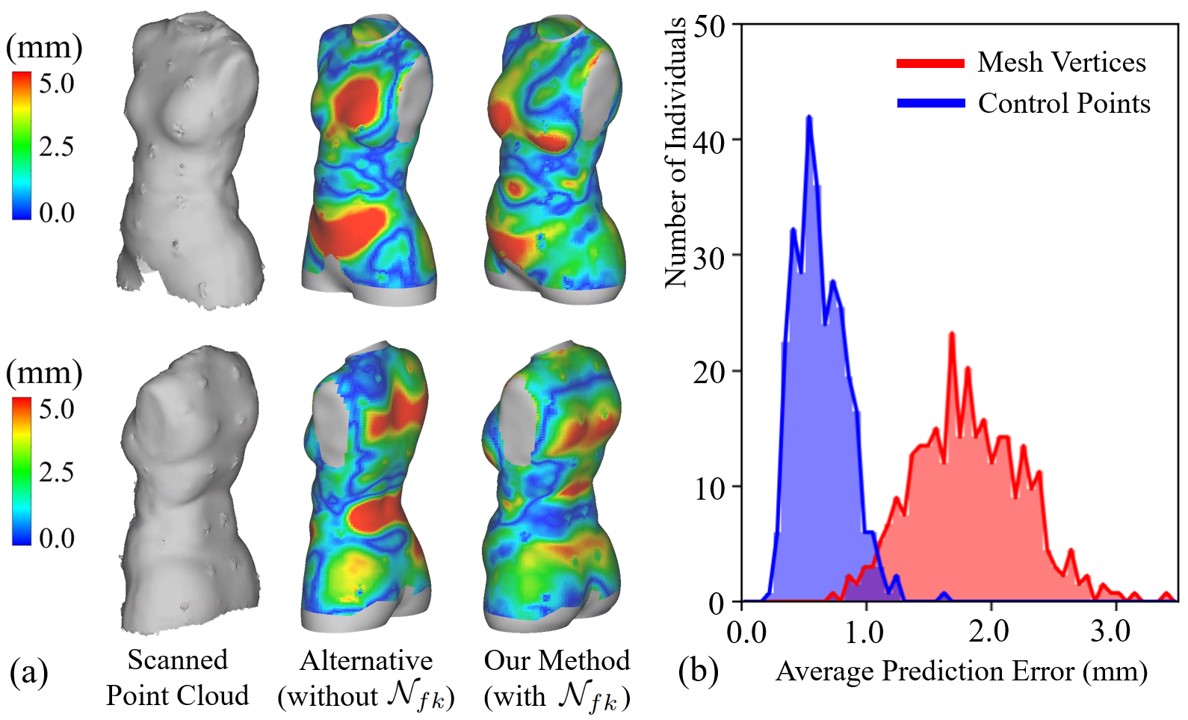}\\
\vspace{-2pt}
\caption{ 
\rev{}{Study of different alternatives. (a) Much larger of shape approximation errors are generated by using an alternative method (without the simulation-based $\mathcal{N}_{fk}$ network) that first predicts the positions of markers from actuation and then deform the body shape by marker-based RBF deformation~\cite{Charlie2007RBFVolPara} -- see results shown in the middle column. (b) Comparison of the shape prediction errors (i.e., the average error of $18,000$ sample points on each model) among a test set with 300 individuals generated by $\mathcal{N}_{fk}$ when outputting the B-spline control points as compact shape descriptor (blue) vs. the mesh vertices (red).} 
}\label{fig:compFKLearning}
\end{figure}

In our implementation, we first train $\mathcal{N}_{fk}$ and then train $\mathcal{N}_{rbf}$ by fixing $\mathcal{N}_{fk}$. The RBF-based space warping function $\mathbf{\Phi}$ and the B-spline shape decoder $\mathbf{B}$ are both implemented as differentiable layers, enabling seamless integration into the standard training process of $\mathcal{N}_{rbf}$. 
\rev{}{Having a surrogate model $\mathcal{N}_{fk}$ that is trained by physics-based simulation is essential to achieve satisfied results of shape prediction. As an alternative, researchers might directly train a network to predict the positions of markers $\{\mathbf{p}^*\}$ by $\mathbf{a}$ and then deform the rest shape of human body by the RBF-based warping function $\Phi(\cdot)$ \cite{Charlie2007RBFVolPara}. However, large shape approximation errors will be generated by this alternative without $\mathcal{N}_{fk}$ (see the middle of Fig.~\ref{fig:compFKLearning}(a)). The other interesting study is about the output of $\mathcal{N}_{fk}$. As can be observed from the comparison given in Fig.~\ref{fig:compFKLearning}(b), the B-spline surface representation used in our pipeline results in much smaller shape approximation error than directly predicting the positions of mesh vertices. The average shape approximate error is evaluated on each model of a test set with 300 individuals, and the distribution of errors among all individuals are plotted as a histogram to conduct the comparison. The mean of error distribution can be reduced from 1.82 mm to 0.63 mm when learning B-spline control points instead of mesh vertices by $\mathcal{N}_{fk}$.
}

\begin{figure*}[t] 
\centering
\includegraphics[width=1.0\linewidth]{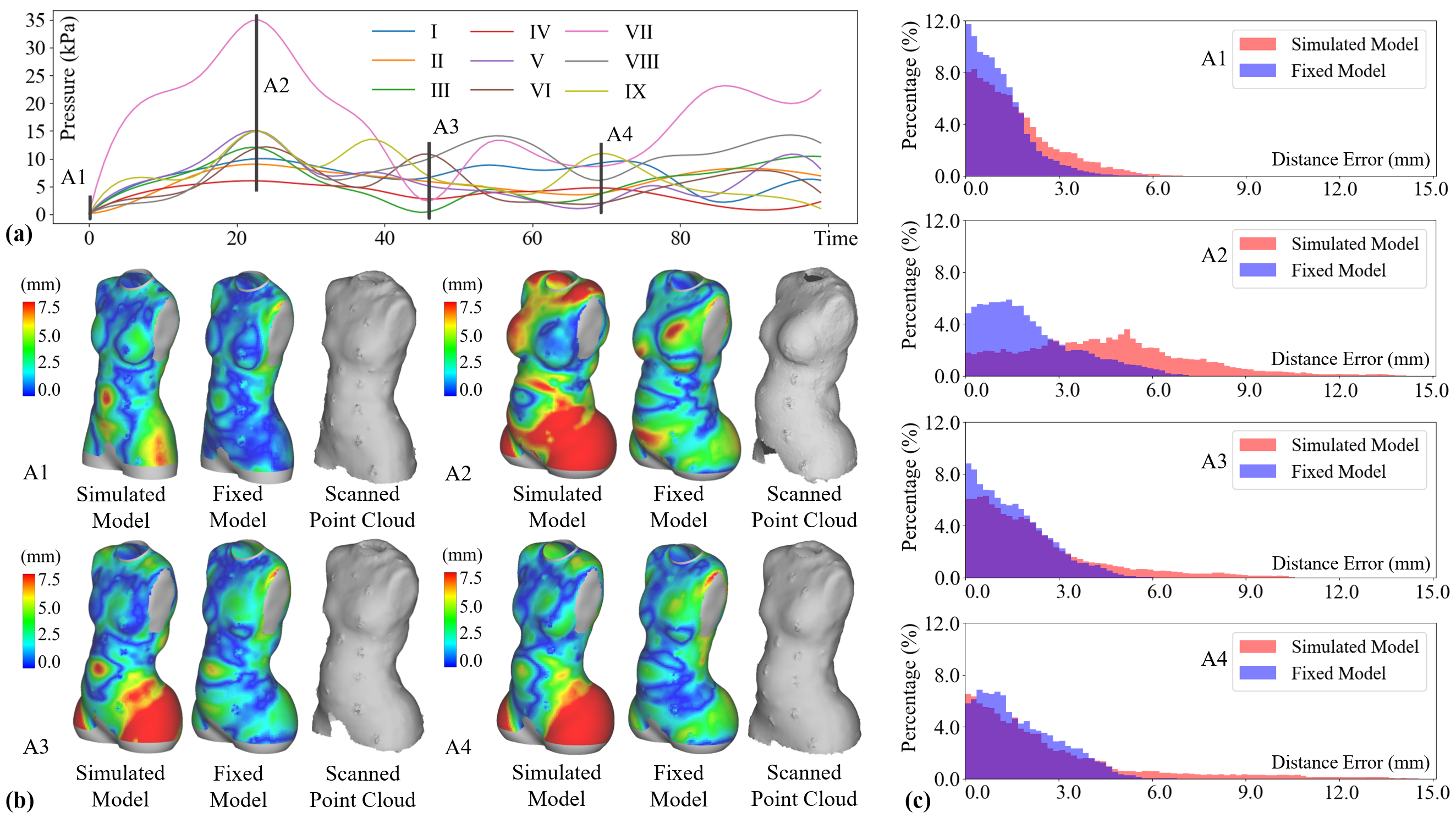}
\caption{Result of different free-form surfaces predicted by our forward kinematics pipeline (i.e., Fig.\ref{fig:NNBasedFK} and Eq.(\ref{eqSim2RealPrediction})) while changing the actuation parameters (a). Four instants (A1-A4) are selected to apply onto the physical setup with the resultant shapes scanned and compared with the predicted shapes, where the shape approximation errors are visualized as colors.
The results with and without sim-to-real transfer (denoted by fixed and simulated models respectively) are compared in both the color maps (b) and the error histograms (c).
}\label{fig:Sim2RealEffectiveStudy}
\end{figure*}

\subsection{Hardware for Verification}\label{subsecResHardware}
A physical setup has been developed to test the effectiveness of our sim-to-real and fast IK methods. The hardware contains a soft robotic mannequin representing both the front and back sides of a human body at a 1:2 scale, a closed-loop pressure control system, and a vision system to capture the positions of markers for training. The deformed shapes of the mannequin are captured by a 3D scanner for verification. 

Our soft robotic mannequin can be deformed by pressurized air and features a design with three layers: an inner soft chamber layer attached to the base solid core (for chambers I-III \& V-IX as shown in Fig.\ref{fig:motivation}(a)); a middle soft chamber layer attached on top of the inner layer (as chamber IV in Fig.\ref{fig:motivation}(a)); and a top membrane layer (passively morphed) covering the base layers to form the overall smooth shape. By pumping air at varying pressures into the chambers, the soft mannequin can be deformed into different shapes. An open-source pneumatic platform, OpenPneu~\cite{tian2023openpneu}, is employed to control the actuation with well-controlled pressures. 

A Vicon motion capture system~\cite{Vicon} with 8 cameras is employed to collect the positions of markers on the deformed surfaces, and a structured-light based 3D scanner Artec Eva Lite~\cite{Scanner} is used to capture the surface geometry of a deformed mannequin as point clouds. Shape errors between the scanned mannequin and the target shape are measured after applying an ICP-based rigid registration to the target shape. Note that as the robotic mannequin is half size, the target shape is also scaled downed to the half size as input in all our experimental tests.

\subsection{Results of Sim-to-Real Learning}\label{subsecResSim2Real}
Experimental tests have been conducted to verify the performance of our sim-to-real method. First of all, we randomly change all the nine actuation parameters of $\mathbf{a}$ and use our forward kinematic network to predict free-form surface shapes. Four instances are selected to apply the actuation $\mathbf{a}$ onto the physical setup. The resultant shapes on the soft robotic mannequin are scanned and compared with the simulated models (i.e., without sim-to-real transfer) and the fixed models (i.e., using the full pipeline with the sim-to-real network). The shape approximate errors are evaluated by first correcting the pose of scanned model using the ICP-based registration and then compute the distances between every surface sample points to their closest points on the scanned model. As can be found from the results shown in Fig.\ref{fig:Sim2RealEffectiveStudy}, the shape approximation errors on the fixed models were significantly reduced. The reductions are $30.4\%$, $53.5\%$, $32.1\%$ and $33.5\%$ for the average error and by $6.7\%$, $44.0\%$, $50.6\%$ and $60.6\%$ for the maximal error. Result can also be found in the supplemental video. 

\begin{figure}[t] 
\centering
\includegraphics[width=1.0\linewidth]{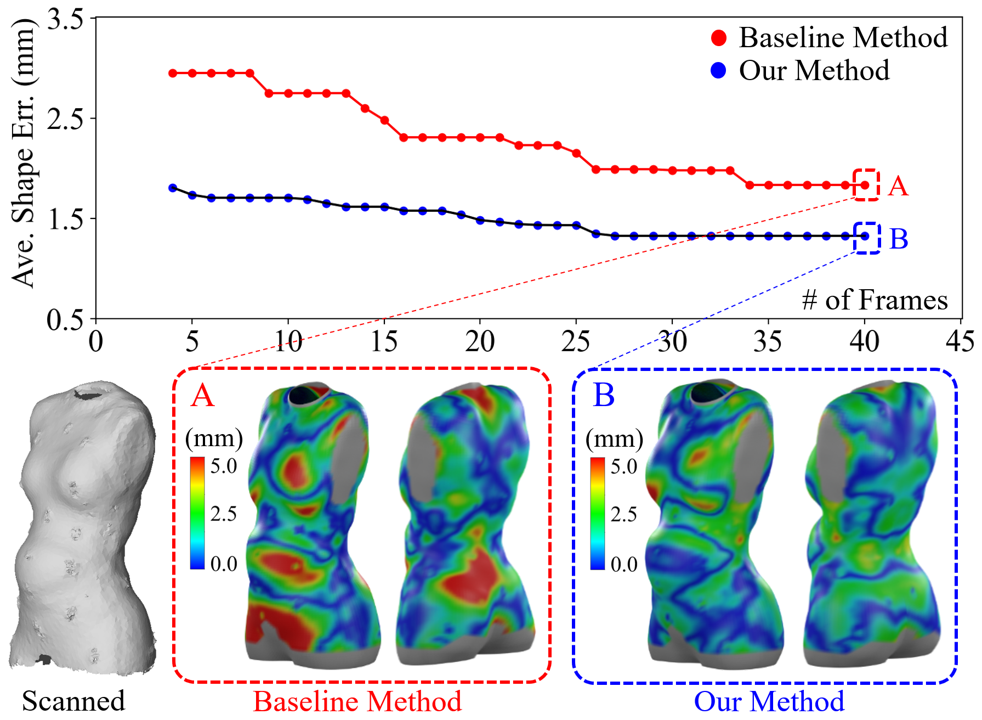}\\
\vspace{-2pt}
\caption{Comparison of our function-prediction based sim-to-real method vs. the marker-prediction based method (as baseline). (Top) The errors between the scanned shape and the predicted shape when more and more frames of markers are employed for the sim-to-real learning. (Bottom) The scanned surface and the error maps of the predicted surfaces.
}\label{fig:Sim2RealComparison}
\end{figure}

We then study the effectiveness of our function-prediction based sim-to-real pipeline by comparing it with the marker-prediction based pipeline as discussed in Sec.\ref{subsecRBFSpaceWarping}, which is used as a baseline. While randomly changing the pneumatic actuation, markers attached on the surface of the physical mannequin are progressively captured by the Vicon system and employed to train the sim-to-real networks. The trained networks are then employed to work together with $\mathcal{N}_{fk}$ to predict the free-form surfaces generated by a random selected actuation $\mathbf{a}_r$. The shape approximation errors are measured between the predicted shapes and the real surface, which is captured by scanning the physical model deformed by applying $\mathbf{a}_r$. The baseline method can only use those frames with full set captured markers for training. As shown in Fig.\ref{fig:Sim2RealComparison}, our method can predict a surface shape with smaller error. This mainly because that all captured markers can contribute to the sim-to-real training of our function-prediction based model even when they are on frames with missing markers. Differently, among all 40 frames involved in the tests shown in Fig.\ref{fig:Sim2RealComparison}, only about 55\% can be employed by the marker-prediction based method. The other 45\% frames all have 1 to 8 missing markers due to the occlusion caused by large deformation on chambers. 

\begin{figure}[t] 
\centering
\includegraphics[width=1.0\linewidth]{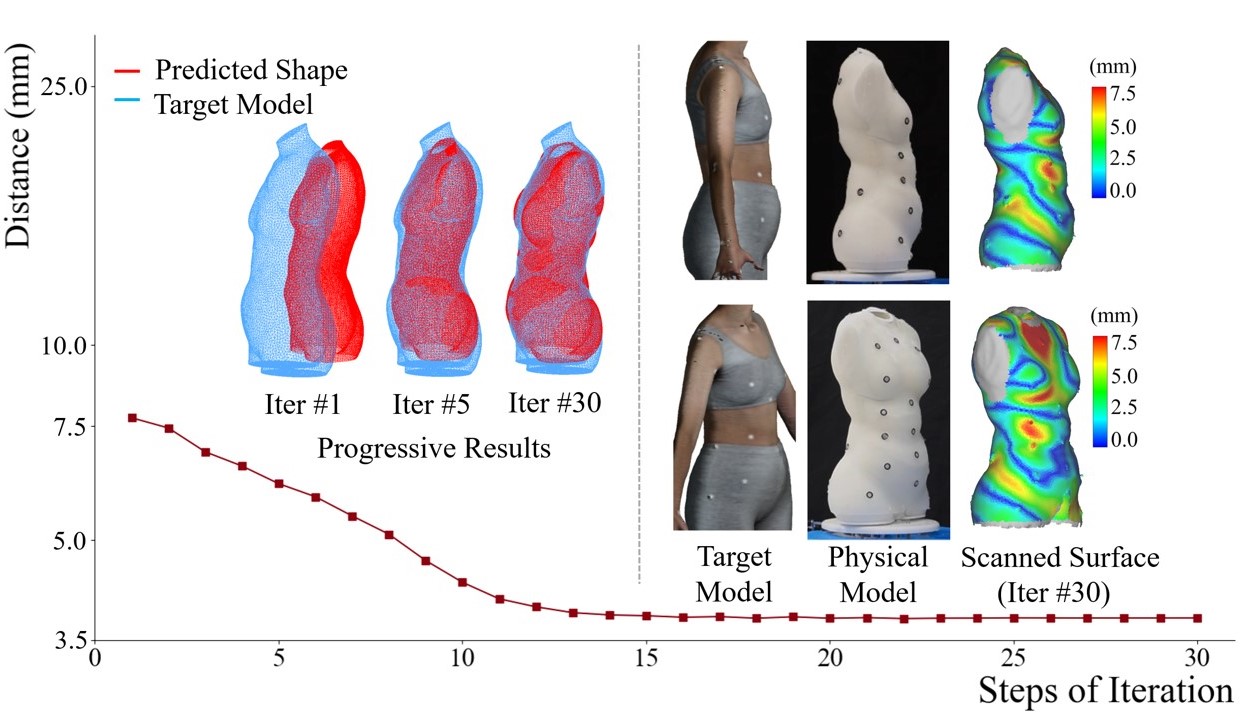}
\caption{The progressive results by applying our NN-based fast IK solver, where the curve shows the average shape approximation errors during the iterations of gradient descent. Using the actuation determined by our fast IK solver, a shape similar to the target model can be realized on the soft mannequin where the error analysis is conducted by 3D scanning with its result shown as the color map.
}\label{fig:progressIKResult}
\end{figure}

\begin{figure}[t] 
\centering
\includegraphics[width=1.0\linewidth]{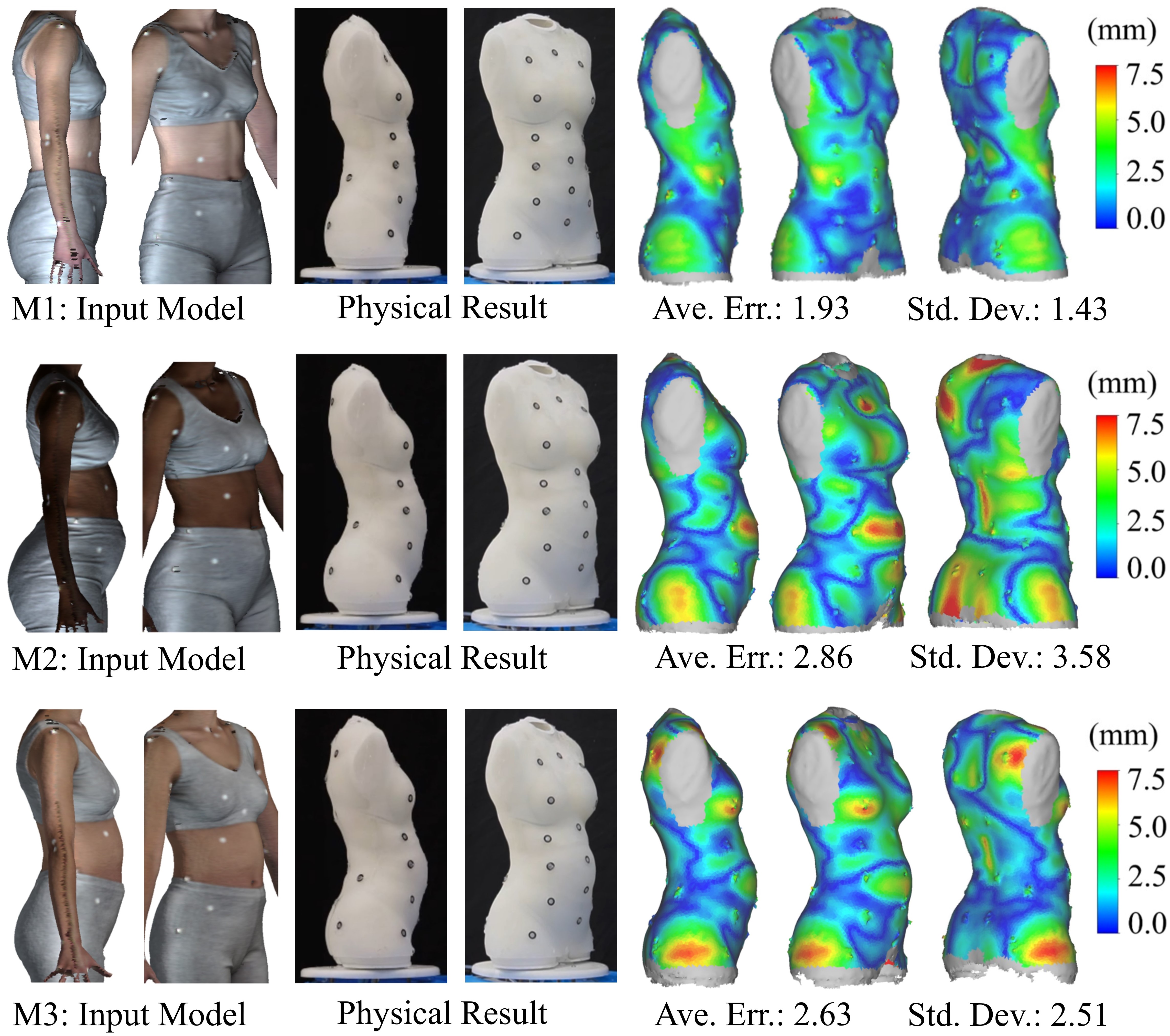}\\
\caption{The physical results of our fast IK solver for three different target models, where the distributions of shape approximation errors are generated by scanning and visualized as the color maps. 
}\label{fig:IKResults}
\end{figure}

\subsection{Results of Inverse Kinematics}\label{subsecResIK}
We have tested the performance of our NN-based fast IK solver on a variety of individual shapes presented in the CAESAR dataset. Figure \ref{fig:progressIKResult} shows the progressive results when applying our IK solver to realize the target shape on the physical setup of pneumatic driven deformable mannequin. It can be observed that our method converges very fast -- i.e., around 15 iterations where each iteration takes less then 0.5 second. 
The distribution of shape approximation errors are generated by scanning the soft mannequin that has been deformed using the resultant actuation determined by our IK solver. 

The effectiveness of our fast IK solver has been tested on other models randomly selected from the CAESAR dataset -- see the results shown in Fig.\ref{fig:IKResults}. It can be find that the shapes realized on the robotic deformable mannequin can well mimic the shape of target models. This can also be verified by the quantitative analysis results generated by the 3D scanner, where the errors are displayed as color maps and reported in the bottom row of Fig.\ref{fig:IKResults}. 

\subsection{\rev{}{Discussion}}


\subsubsection{\rev{}{Computational efficiency}}
In this work, the NN-based IK computing pipeline enables fast and robust shape approximation for a deformable free-form surface. We have achieved a computing speed of less than 10 seconds per target model, which is significantly faster than existing methods and aligns well with the physical actuation time of the proposed setup. 
\rev{}{The computational complexity of our pipeline is influenced by various factors, including the network architecture, the number of control points on the B-spline surface, and the kernel number of the RBF morphing function. The current selection is mainly based on experimental tests to find the balance between performance and efficiency. We can further accelerate the computation by GPU-based parallel computing.}

\subsubsection{\rev{}{Applicability to other deformable robots}}
\rev{}{Our pipeline is also applicable to other soft and deformable robotic systems. In principle, our function based sim-to-real learning method can be applied to a variety of deformable robots when their freeform shapes can be represented by B-spline surfaces (e.g., Chen \textit{et al.}~\cite{chen2023morphological}). We will further explore the potential of our method in other robotic setups, particularly for addressing data collection challenges with learning-based solutions.}

\subsubsection{\rev{}{Different setups for data acquisition}}
In this work, a well-calibrated motion capture system (with position tracking error less than 0.1 mm \cite{viconAccuracy}) is applied to capture data from the physical setup. 
\rev{}{While our pipeline is designed for sparse yet accurate data from motion capture, it has the potential to be extended to cover the systems with LiDAR~\cite{LidarCamIntel} or RGBD cameras~\cite{IntelD457Camera, kurillo2022evaluating} that capture dense but less accurate point clouds.}
\rev{}{On the other hand, since the material of the deformable surface exhibits hysteresis,} the positions of markers are captured in a quasi-static manner. As such, we \rev{always}{have to} wait \rev{}{for} 3 seconds after applying the actuation to make the system stable and then capture frames of markers. \rev{}{This represents a limitation of the proposed method when applied to robotic systems that require high dynamic performance. One possible solution is incorporating other techniques that can handle sequential data inputs in the learning pipeline (e.g.,~\cite{Thuruthel2017DynamicsLearning}).}

\subsubsection{\rev{}{Design of deformable mannequin}}
The design of the proposed deformable surface can be further enhanced. \rev{}{This includes optimizing the shape and location of pneumatic-driven chambers, as well as the geometry of the rigid core.} In our current design, we employed a neural distance field based method~\cite{DeepSDF2019} to determine a volumetric region that is commonly `inside' the body shapes of all individuals. \rev{}{We have not considered the poses of individuals when determining the shape of the core. We plan to address this limitation in our future work.}

\section{Conclusion}

In this paper, we addressed the challenge of bridging the gap between simulated and physically deformed free-form surfaces caused by hardware errors and simplifications in physical simulation. Our proposed novel approach, based on deformation function-driven sim-to-real learning, offers a promising solution. It effectively maps simulated model geometries to their real-world counterparts, even accommodating sparsely distributed set of markers with missing data. By integrating this method into a neural network-based computational pipeline, we demonstrate its effectiveness in solving the inverse kinematic problem for deformable mannequins with pneumatic actuation. Through this work, we contribute to advancing the field of shape control for deformable free-form surfaces, paving the way for more accurate and reliable realization in various applications. 

\section*{Acknowledgments}
The project is partially supported by the chair professorship fund of the University of Manchester and the research fund of UK Engineering and Physical Sciences Research Council (EPSRC) (Ref.\#: EP/W024985/1).

\bibliographystyle{plainnat}
\bibliography{references}

\end{document}